\documentclass[sigconf,screen]{acmart}
\AtBeginDocument{%
  }

\setcopyright{acmlicensed}
\copyrightyear{2026}
\acmYear{2026}
\acmDOI{XXXXXXX.XXXXXXX}
\acmConference[MM '26]{the 34rd ACM International Conference on Multimedia}{November 10--14,
  2026}{Rio de Janeiro, Brazil}
\usepackage{cleveref}
\crefname{table}{Table}{Tables}
\crefname{figure}{Figure}{Figures}
\usepackage{colortbl}
\usepackage{array}
\usepackage{amsmath, bm}
\usepackage{multirow}
\usepackage{makecell}
\usepackage{graphicx}
\usepackage{algorithmicx,algorithm}
\usepackage[dvipsnames,svgnames]{xcolor}
\usepackage{xcolor}
\usepackage{svg}
\usepackage{alltt}
\usepackage{caption}
\usepackage{wrapfig}
\usepackage{pifont}       % \ding{xx}
\usepackage{bbding}       % \Checkmark,\XSolid,... (需要和pifont宏包共同使用)
\usepackage{fontawesome}  % \faCheck,\faTimes
\usepackage{tikz}
\usepackage{booktabs}
\usepackage{hyperref}
\usepackage{threeparttable}

\definecolor{pink}{HTML}{DAA2B6}
\definecolor{brownnew}{HTML}{AA5757}
\definecolor{greennew}{HTML}{91CD68} 
\definecolor{bluenew}{HTML}{5EA8ED}
\definecolor{orangenew}{HTML}{E6AC83}

\usepackage[table]{xcolor} % 支持表格上色
\definecolor{lightblue}{RGB}{235, 245, 255} % 浅蓝色 (RGB)
\definecolor{pubblue}{RGB}{50, 50, 250} % 紫色

\setcopyright{acmlicensed}
\copyrightyear{2026}
\acmYear{2026}
\acmDOI{XXXXXXX.XXXXXXX}% 录用后再填
\acmISBN{978-1-4503-XXXX-X/2018/06}
\acmSubmissionID{2116}   % 替换成你的 Submission ID

\begin{document}

\title{Enhancing Self-Supervised Talking Head Forgery Detection via a Training-Free Dual-System Framework}

\author{Ke Liu}
\affiliation{%
  \institution{University of Electronic Science \\ and Technology of China}
  \city{Chengdu}
  \country{China}
}
% \email{liuke3068@gmail.com}

\author{Jiwei Wei}
\authornote{Corresponding Author}
\affiliation{%
  \institution{University of Electronic Science \\ and Technology of China}
  \city{Chengdu}
  \country{China}
}
% \email{liuke3068@gmail.com}
\author{Shuchang Zhou}
\affiliation{%
  \institution{University of Electronic Science \\ and Technology of China}
  \city{Chengdu}
  \country{China}
}

\author{Yutong Xiao}
\affiliation{%
  \institution{University of Electronic Science \\ and Technology of China}
  \city{Chengdu}
  \country{China}
}

\author{Ruikun Chai}
\affiliation{%
  \institution{University of Electronic Science \\ and Technology of China}
  \city{Chengdu}
  \country{China}
}

\author{Yitong Qin}
\affiliation{%
  \institution{University of Electronic Science \\ and Technology of China}
  \city{Chengdu}
  \country{China}
}

\author{Yuyang Zhou}
\affiliation{%
  \institution{Hainan University}
  \city{Haikou}
  \country{China}
}

\author{Yang Yang}
\affiliation{%
  \institution{University of Electronic Science \\ and Technology of China}
  \city{Chengdu}
  \country{China}}
  
% \affiliation{%

% \affiliation{%

% \affiliation{%

% \affiliation{%

% \affiliation{%

% \affiliation{%

% \affiliation{%


\begin{abstract}
Supervised talking head forgery detection faces severe generalization challenges due to the continuous evolution of generators. By reducing reliance on generator-specific forgery patterns, self-supervised detectors offer stronger cross-generator robustness. However, existing research has mainly focused on building stronger detectors, while the discriminative capacity of trained detectors remains insufficiently exploited. In particular, for score-based self-supervised detectors, the limited discriminative ability on hard cases is often reflected in unreliable anomaly ordering, leaving room for further refinement. Motivated by this observation, we draw inspiration from the dual-system theory of human cognition and propose a Training-Free Dual-System (TFDS) framework to further exploit the latent discriminative capacity of existing score-based self-supervised detectors. TFDS treats anomaly-like scores as the basis of System-1, using lightweight threshold-based routing to partition samples into confident and uncertain subsets. System-2 then revisits only the uncertain subset, performing fine-grained evidence-guided reasoning to refine the relative ordering of ambiguous samples within the original score distribution. Extensive experiments demonstrate consistent improvements across datasets and perturbation settings, with the gains arising mainly from corrected ordering within the uncertain subset. These findings show that existing self-supervised talking head forgery detectors still contain underexploited discriminative cues that can be effectively unlocked through training-free dual-system reasoning.
\end{abstract}






\maketitle

\section{Introduction}
Recent advances in generative models have made talking head forgeries increasingly realistic \cite{peng2024synctalk,dong2025talking}, substantially lowering the barrier to the large-scale creation and dissemination of harmful AI-generated content \cite{yu2024gaussiantalker}. Reliable talking head forgery detection has therefore become increasingly important. The main difficulty lies in maintaining robust generalization to unseen generators, rather than merely recognizing known manipulations \cite{guo2025towards}. This challenge is particularly severe for supervised detectors, as their discriminative capacity is often tied to generator-specific signatures.

\begin{figure}[t]
\centering
\includegraphics[scale=0.44]{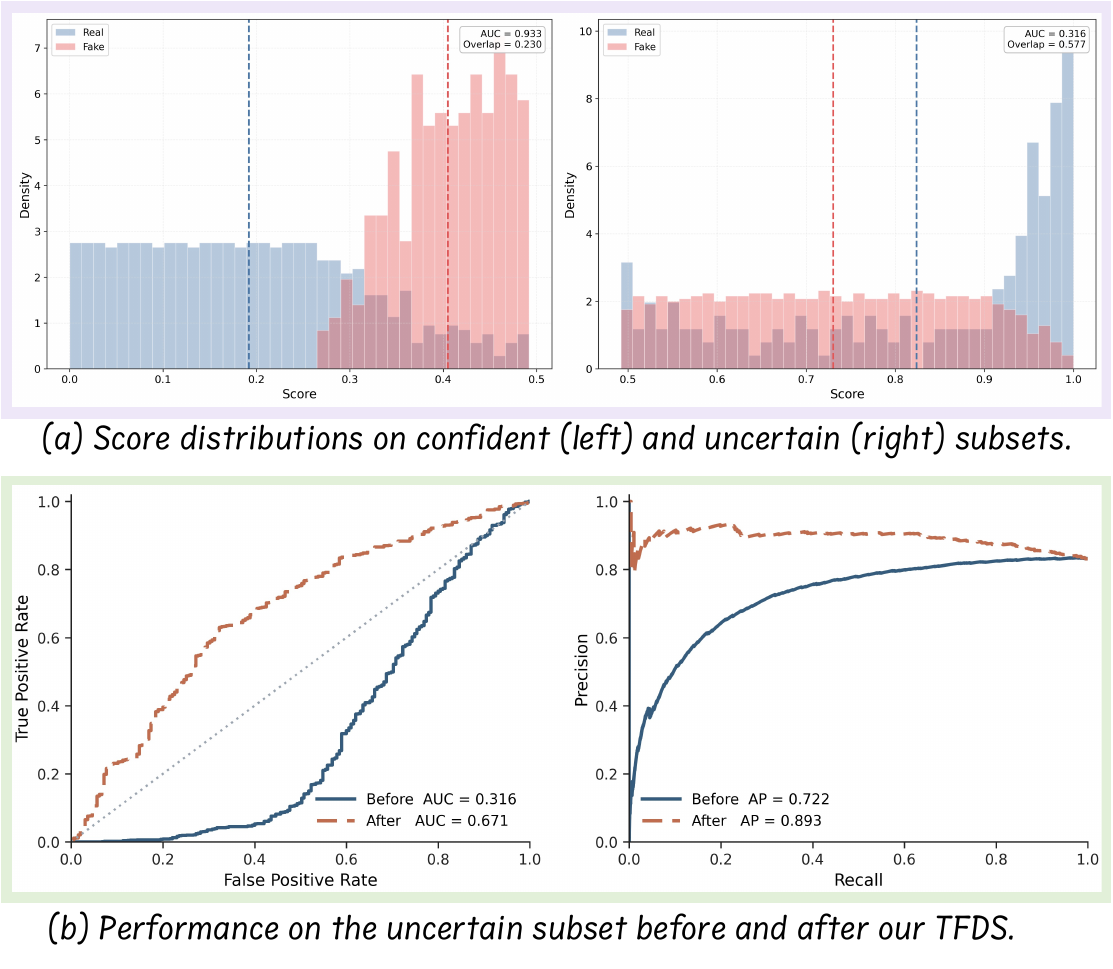}
\caption{The main detection difficulty is concentrated on the uncertain subset, and TFDS improves hard-sample separability. (a) Although the existing self-supervised detector separates confident samples reliably, the overlap between real and fake samples becomes substantially heavier on the uncertain subset, thereby limiting overall detection performance. (b) After introducing TFDS, the ROC and PR curves on the uncertain subset improve markedly, indicating that TFDS effectively enhances hard-sample separability.}
\label{fig1}
% \vspace{-0.5em}
\end{figure} 

Compared with supervised detectors, self-supervised methods \cite{smeu2025circumventing,feng2023self} are inherently better suited to robust generalization, as they rely primarily on authenticity-related patterns. Such methods typically formulate detection as anomaly detection \cite{liu2023learning}, treating samples that deviate from the real distribution as forgeries. However, as forged talking head videos increasingly approach the real distribution, the deviation signals on which detection relies become progressively compressed, making reliable discrimination substantially more difficult. This limitation is concentrated mainly on hard samples. As shown in Figure \ref{fig1}(a), the detector \cite{smeu2025circumventing} can separate confident samples reliably, while its discriminative ability becomes limited on the uncertain subset. Unlike existing efforts \cite{liu2024lips,kukanov2025klassify}, which respond by designing stronger detectors, we instead ask whether the discriminative capacity already encoded in an existing detector can be further exploited.

Based on the above observations, we find that the differentiated behavior of the score-based self-supervised detector on confident and uncertain samples motivates a dual-system perspective \cite{zhang2025system}. In human cognition, System-1 typically produces fast judgments, while ambiguous or easily confusable cases are left for System-2 to analyze more deliberately. Viewed in this way, the self-supervised detector can be regarded as a System-1 mechanism, effective at handling easy cases yet insufficient for discrimination on boundary cases. Since such detection is fundamentally score-based, this limitation is concretely reflected in unreliable score ordering among hard samples. This shifts the focus from designing a stronger detector to introducing System-2 to revisit uncertain samples. In this regard, Multimodal Large Language Models (MLLMs) are particularly attractive because of their strength in fine-grained evidence analysis.

In recent years, MLLMs have emerged as a promising direction for forgery detection \cite{gu2025allm4add,huang2025sida}, owing to their strong cross-modal modeling capability \cite{yang2026endowing,wei2023less}. They can provide richer semantic evidence and more interpretable analytical cues, opening new possibilities for fine-grained discrimination in complex forgery scenarios. However, most existing approaches still integrate MLLM capabilities into the detector training pipeline \cite{yu2025unlocking}. Such training-coupled designs inevitably increase training cost and deployment complexity, and remain ill-suited to the rapid evolution of forgery paradigms. In contrast, we do not use large models as detector components that require further training or redesign. Instead, we use them as a training-free System-2 that is activated only for uncertain samples, thereby exploiting their strength in fine-grained reasoning precisely where it is most needed. In Figure \ref{fig1}(b), this design enables more effective discrimination within the uncertain subset.

Building on the above analysis, we develop a Training-Free Dual-System (TFDS) framework for talking head forgery detection. Built on a self-supervised detector \cite{smeu2025circumventing}, System-1 estimates a Youden threshold \cite{fluss2005estimation} on the validation set and partitions test samples into confident and uncertain subsets through lightweight score-based routing. Confident samples retain the original predictions, while uncertain samples are forwarded to System-2. Since audio-visual consistency has already been modeled by the base detector, System-2 is introduced not to repeat multimodal alignment modeling, but to provide a complementary vision-language perspective for revisiting uncertain samples. Specifically, it uses frozen CLIP \cite{radford2021learning} and predefined text prototypes to score video frames and local patches, identify salient evidence regions, and construct cross-frame patch strips as localized evidence. Qwen \cite{wang2024qwen2} then performs vision-language reasoning on this evidence to produce fine-grained descriptions, which are converted into ranking scores by a text reranker \cite{rachidy2025domain}. The resulting refinement is restricted to the original score distribution and only adjusts local ordering within the uncertain subset, thereby preserving the global decision structure of the base detector. In this way, TFDS preserves the generalization strength of the original detector while further unlocking its latent discriminative capacity on uncertain cases, with experimental results showing that its gains mainly come from improved ordering of hard samples.

In summary, our main contributions are as follows:
\begin{itemize}
\item We present a new perspective on talking head forgery detection by showing that the main bottleneck of existing self-supervised detectors lies in insufficient exploitation of hard cases, rather than only in the need to design ever stronger detectors.

\item We propose TFDS, a training-free dual-system framework that builds System-1 on lightweight score-based routing and introduces System-2 for fine-grained evidence-guided reasoning, while restricting refinement to local reordering within the original score distribution.

\item Extensive experiments on multiple benchmarks demonstrate that TFDS consistently improves the base detector, with gains concentrated on uncertain samples and remaining robust under diverse perturbation settings.
\end{itemize}
% \begin{itemize}

% \item We provide a new perspective on talking head forgery detection by highlighting the underexploited discriminative capacity of an existing self-supervised detector on uncertain cases, offering an alternative to repeatedly redesigning stronger detectors. 
% \item We propose TFDS, a training-free dual-system framework that couples lightweight score-based routing in System-1 with large model-based fine-grained reasoning in System-2, while constraining refinement within the original score distribution. 
% \item Extensive experiments across multiple datasets show that TFDS consistently improves performance over the original detector, with the gains remaining robust under diverse perturbation settings.
% \end{itemize}

\section{Related Work}
\subsection{Talking Head Forgery Detection}
Talking head generation synthesizes temporally aligned facial motions from speech by modeling cross-modal audio-visual interactions \cite{chencafe2025iclr, wei2023less}. As the realism of such videos improves, corresponding detection methods have evolved from unimodal forgery analysis \cite{huang2023implicit,zheng2021exploring} to audio-visual collaborative modeling \cite{liu2024lips, yang2023avoid}. Most existing methods address this challenge by developing increasingly stronger detectors, often with improved multimodal modeling capability. Early methods directly learn audio-visual detectors from labeled real and forged videos \cite{chugh2020not,mittal2020emotions}, while later approaches combine self-supervised audio-visual representation learning with labeled adaptation \cite{zeng2021contrastive,haliassos2022leveraging, zhou2021joint}. Although effective in-domain, such methods often struggle to cope with generator shift.

To improve generalization, recent studies have further explored anomaly-based detection strategies trained only on real data \cite{li2024zero,ricker2024aeroblade}. AVAD \cite{feng2023self} models temporal synchronization between video and audio through autoregressive learning, while AVH-Align \cite{smeu2025circumventing} improves robustness by leveraging self-supervised audio-visual representations and reducing dataset-specific biases. Compared with fully supervised detectors, these methods are better suited to cross-generator generalization.

However, as talking head generators evolve rapidly, the discriminative signals exploited by anomaly-based detectors become progressively compressed, leaving uncertain samples insufficiently resolved. Rather than continuing to address this issue by retraining a stronger detector, our work instead asks whether an existing self-supervised detector still contains underexploited discriminative potential on uncertain cases.

\subsection{Vision-Language Reasoning}
In recent years, growing efforts have explored the use of Large Language Models (LLMs) and Multimodal Large Language Models (MLLMs) to enhance vision-language reasoning, particularly in fine-grained recognition \cite{li2024llms} and multimodal understanding tasks \cite{oh2025understanding}. One line of work improves model adaptability by introducing additional learnable components, including prompt tuning \cite{zhou2022learning, qi2025global} and lightweight adapters \cite{gao2024clip}. Although effective, these approaches still require trainable modules and often depend on labeled data or additional optimization. To reduce such overhead, cache-based methods such as Tip-Adapter \cite{zhang2022tip} and its unsupervised variants TDA \cite{karmanov2024efficient} and DMN \cite{zhang2024dual} perform adaptation through reference-feature retrieval. However, their performance remains highly dependent on the quality of cached samples, making them less stable in data-scarce or noisy settings.

Another related line of research enhances visual understanding through text-based reasoning with LLMs. DCLIP \cite{menon2022visual} uses GPT-3 \cite{brown2020language} to enrich category names with attribute-level descriptions. HIE \cite{ren2023chatgpt} introduces hierarchical discriminative descriptions for category reasoning. CuPL \cite{pratt2023does} replaces manually designed prompts with LLM-generated ones, and ProAPO \cite{qu2025proapo} further scales prompt generation and optimization. Despite their differences, these methods improve vision-language modeling either by adapting pretrained models with lightweight mechanisms or by enriching textual reasoning with external knowledge. 

These studies inspire us to transfer fine-grained vision-language reasoning to the hard samples that remain unresolved by the self-supervised talking head forgery detector.

\subsection{Training-Free Forgery Detection}
In recent years, training-free detection \cite{wang2024lampmark} has emerged as an alternative to training-based forgery detectors. Instead of learning a new detector, these methods directly exploit the representations or statistical properties already encoded in pretrained foundation models to distinguish real from fake samples. AeroBlade \cite{ricker2024aeroblade} uses reconstruction errors in latent diffusion models, RIGID \cite{he2024rigid} exploits differences in sensitivity to random perturbations, and ZED \cite{cozzolino2024zero} adapts likelihood-based ideas from AI-generated text detection within a multi-level super-resolution framework. These studies show that effective forgery cues can be derived from reconstructability or statistical consistency without retraining a dedicated detector. More recent work further extends this paradigm. The role of foundation-model robustness in training-free detection has been analyzed in \cite{tsai2024understanding}. WaRPAD \cite{choi2025training} constructs a training-free detector from cropping robustness and local patch scoring. 

However, these methods are mainly developed for fake image detection, and training-free paradigms remain underexplored in talking head forgery detection. More importantly, existing training-free methods typically construct a new detection score from raw inputs, whereas our method builds on an existing self-supervised audio-visual detector and exploits its latent discriminative potential on uncertain samples through a dual-system framework.

\section{Method}
\subsection{Overview}
Given a test set of videos, let $s_i$ denote the original score produced by an existing self-supervised audio-visual detector \cite{smeu2025circumventing} for a sample $x_i$, where a larger $s_i$ indicates that the sample is more likely to be fake. Based on the validation set, a routing threshold $\tau$ is estimated using the Youden criterion \cite{fluss2005estimation}, which is then used to partition test samples into a confident subset $\mathcal{C}$ and an uncertain subset $\mathcal{U}$. For each $x_i \in \mathcal{U}$, System-2 first mines localized suspicious evidence with a frozen CLIP model \cite{radford2021learning} and predefined real/fake text prototypes generated by GPT-4 \cite{sanderson2023gpt}, then generates fine-grained descriptions with Qwen \cite{wang2024qwen2}, and finally produces a score $r_i$ through a text reranker \cite{rachidy2025domain}. TFDS keeps the scores of $\mathcal{C}$ unchanged and only refines samples in $\mathcal{U}$ by locally reordering them according to $r_i$, after which they are reassigned to the original score slots $\mathcal{V}_{\mathcal{U}}$ to obtain the final score $\hat{s}_i$.

\begin{figure}[t]
\centering
\includegraphics[scale=0.5]{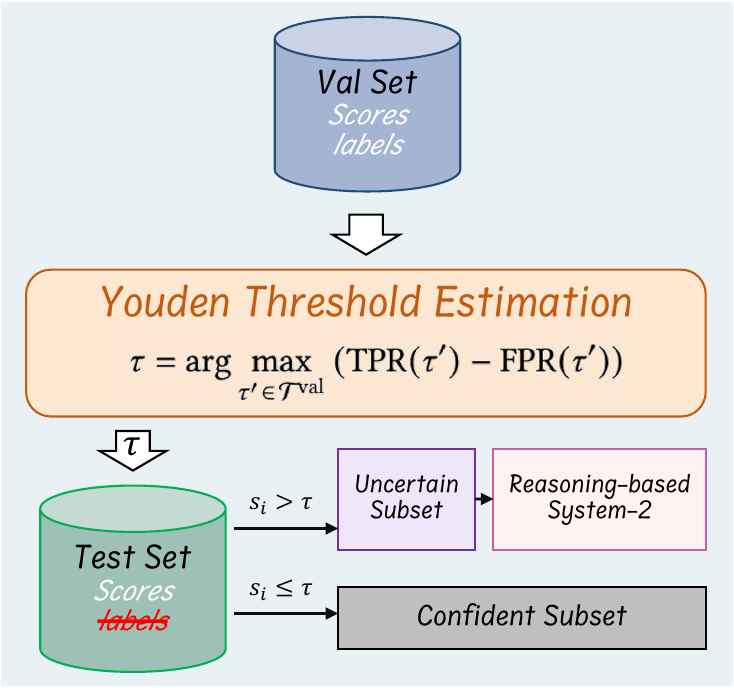}
\caption{System-1 for uncertainty routing. Labels are used only on the validation set to estimate the threshold $\tau$ with the Youden criterion. Test samples are partitioned into the confident subset $\mathcal{C}$ and uncertain subset $\mathcal{U}$ by thresholding detector scores with $\tau$, without using any test labels. Confident samples retain the original detector predictions, while uncertain samples are forwarded to System-2.}
\label{fig2}
% \vspace{-0.5em}
\end{figure}
\subsection{System-1: Uncertainty Routing}
As illustrated in Figure \ref{fig2}, System-1 is built on top of an existing self-supervised audio-visual detector \cite{smeu2025circumventing} and serves as a lightweight routing module. Its role is not to replace the original detector with a newly trained model, but to preserve the detector’s global decision structure while explicitly isolating the samples that remain insufficiently resolved. For a test video $x_i$, let $s_i$ denote the original score produced by the detector $f(\cdot)$, where a larger $s_i$ indicates that $x_i$ is more likely to be fake. 

System-1 inherits the original detector’s ability to model audio-visual consistency, while exposing the subset of samples on which additional reasoning is still needed. To identify such samples, we estimate a routing threshold $\tau$ on the labeled validation set $\mathcal{D}^{\mathrm{val}}$. Let $\mathcal{T}^{\mathrm{val}}$ denote the set of candidate thresholds induced by validation scores. $\tau$ is selected using the Youden criterion:
\begin{equation}
\tau = \arg\max_{\tau' \in \mathcal{T}^{\mathrm{val}}} \left(\operatorname{TPR}(\tau')-\operatorname{FPR}(\tau')\right).
\end{equation}

\begin{figure*}[t]
\centering
\includegraphics[scale=0.56]{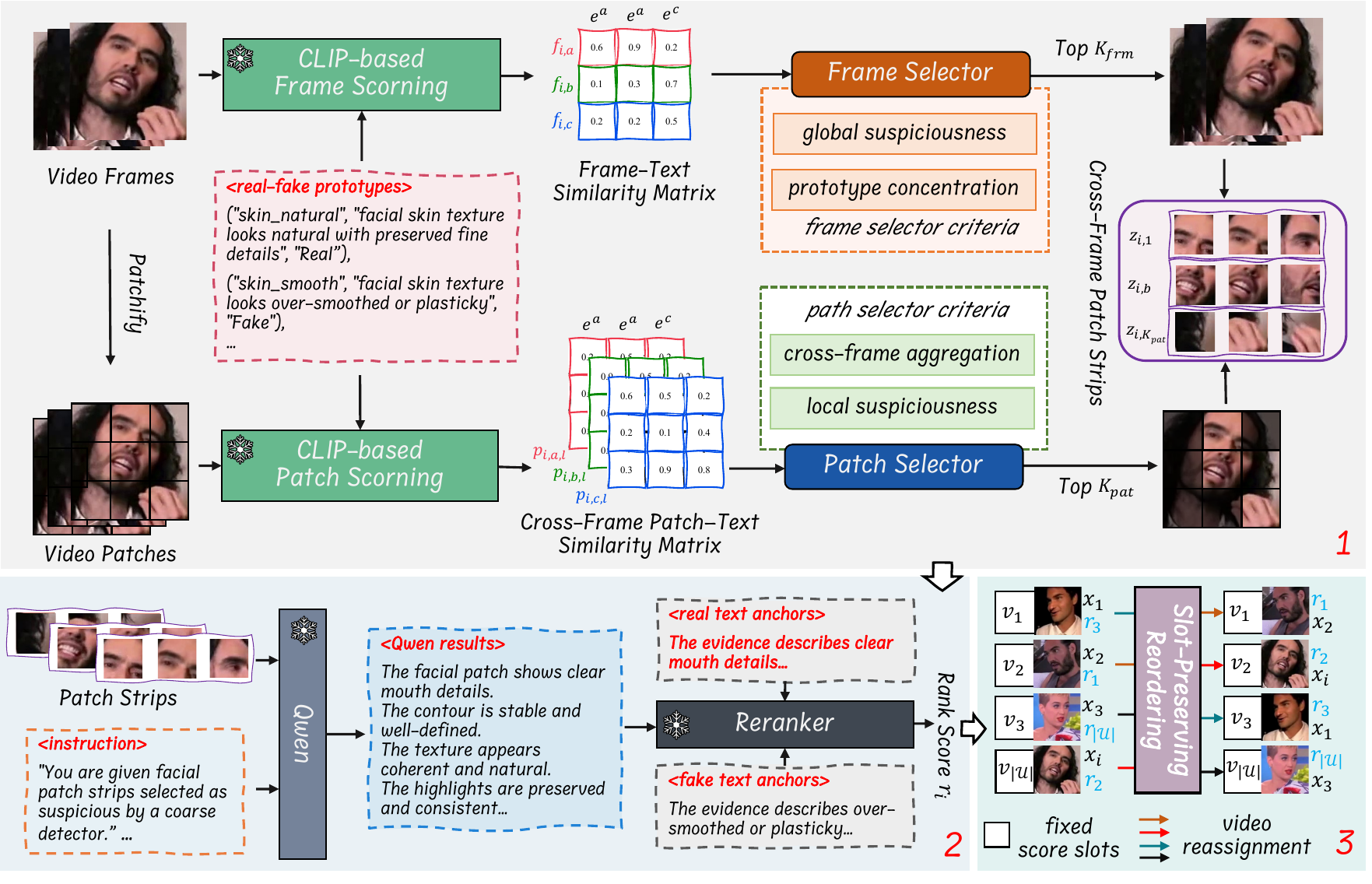}
\caption{Overview of System-2 for fine-grained evidence-guided reasoning and slot-preserving refinement. (1) For an uncertain video, frozen CLIP scores sampled frames and local patches with predefined real/fake text prototypes. Frame selection is guided by global suspiciousness and prototype concentration, while patch selection is guided by local suspiciousness and cross-frame aggregation. The selected top-$K_{\mathrm{frm}}$ frames and top-$K_{\mathrm{pat}}$ patch locations are then organized into cross-frame patch strips. (2) The patch strips, together with a fixed instruction prompt, are fed into Qwen to generate fine-grained evidence descriptions, which are further compared against real and fake text anchors by a reranker to produce a rank score $r_i$. (3) The rank scores are used only to reorder uncertain samples within the fixed original score slots, yielding slot-preserving refinement without redefining the global score distribution.}
\label{fig3}
% \vspace{-0.5em}
\end{figure*}

We adopt the Youden criterion because it provides a simple and stable way to balance True Positive Rate (TPR) and False Positive Rate (FPR) under a single threshold. This is particularly suitable for our purpose, since System-1 is intended to function as a routing mechanism rather than a separately optimized classifier. Instead of introducing additional calibration procedures or more complex threshold learning, the Youden criterion directly determines a validation-based operating point that separates samples already handled reliably by the detector from those that remain ambiguous. Once $\tau$ is obtained, test samples are partitioned into a confident subset $\mathcal{C}$ and an uncertain subset $\mathcal{U}$ according to:
\begin{equation}
\mathcal{C}=\{x_i \mid s_i \le \tau\}, \qquad
\mathcal{U}=\{x_i \mid s_i > \tau\}.
\end{equation}

Samples in $\mathcal{C}$ retain the original detector scores, whereas samples in $\mathcal{U}$ are forwarded to System-2 for further analysis. Rather than retraining a stronger detector to overwrite the original decision process, System-1 first performs efficient global screening and then explicitly exposes the unresolved hard cases. As a result, the discriminative strengths of the existing self-supervised detector are preserved, while the scope of subsequent fine-grained reasoning is restricted to the uncertain subset where it is most needed.

\subsection{System-2: Evidence-Guided Reasoning}
\subsubsection{Fine-Grained Evidence Mining and Rank Score Estimation}

As shown in Figure \ref{fig3}(1) and (2), System-2 is activated only for samples in the uncertain subset $\mathcal{U}$. The base self-supervised detector has already provided an assessment of these samples through audio-visual consistency modeling. System-2 is then introduced to resolve the remaining ambiguity through fine-grained vision-language reasoning.

For an uncertain video $x_i \in \mathcal{U}$, we first construct a frame sequence $\mathcal{F}_i=\{f_{i,t}\}_{t=1}^{T_i}$ by selecting frames according to grayscale frame differences, so as to retain visually informative temporal changes. Each sampled frame is further divided into local patches, where $p_{i,t,\ell}$ denotes the $\ell$-th patch of frame $f_{i,t}$. To mine suspicious evidence, we employ a frozen CLIP encoder $\phi(\cdot)$ together with predefined real and fake text prototype sets $\mathcal{E}^{r}$ and $\mathcal{E}^{f}$. Here, $\mathcal{E}^{r}$ and $\mathcal{E}^{f}$ collect textual prototypes describing real-related and fake-related facial evidence, respectively. 
% CLIP is used here not as a detector, but as an evidence miner that measures how strongly a frame or a patch aligns with these textual priors. 
For frame selection, we use two complementary criteria. The first is \emph{global suspiciousness}, which measures the extent to which frame $f_{i,t}$ favors fake-related prototypes over real-related ones:
\begin{equation}
g^{\mathrm{frm}}_{i,t}
=
\max_{\mathbf{e}^{f}\in\mathcal{E}^{f}} \operatorname{sim}\!\left(\phi(f_{i,t}),\mathbf{e}^{f}\right)
-
\max_{\mathbf{e}^{r}\in\mathcal{E}^{r}} \operatorname{sim}\!\left(\phi(f_{i,t}),\mathbf{e}^{r}\right).
\end{equation}
The second is \emph{prototype concentration}, which measures whether the fake-side response is dominated by a small number of fake prototypes rather than being diffusely distributed over the entire fake prototype set:
\begin{equation}
c^{\mathrm{frm}}_{i,t}
=
\max_{\mathbf{e}^{f}\in\mathcal{E}^{f}} \operatorname{sim}\!\left(\phi(f_{i,t}),\mathbf{e}^{f}\right)
-
\frac{1}{|\mathcal{E}^{f}|}
\sum_{\mathbf{e}^{f}\in\mathcal{E}^{f}}
\operatorname{sim}\!\left(\phi(f_{i,t}),\mathbf{e}^{f}\right).
\end{equation}
A larger $c^{\mathrm{frm}}_{i,t}$ indicates that the strongest fake-side response is more dominant relative to the average response over the fake prototype set. The final frame score is then defined as:
\begin{equation}
u^{\mathrm{frm}}_{i,t}
=
g^{\mathrm{frm}}_{i,t}
+
c^{\mathrm{frm}}_{i,t}.
\end{equation}
In this way, selected frames are required to be both globally suspicious and semantically focused.

For patch selection, we instead use \emph{local suspiciousness} and \emph{cross-frame aggregation}. The local suspiciousness of patch $p_{i,t,\ell}$ is defined as:
\begin{equation}
g^{\mathrm{pat}}_{i,t,\ell}
=
\max_{\mathbf{e}^{f}\in\mathcal{E}^{f}} \operatorname{sim}\!\left(\phi(p_{i,t,\ell}),\mathbf{e}^{f}\right)
-
\max_{\mathbf{e}^{r}\in\mathcal{E}^{r}} \operatorname{sim}\!\left(\phi(p_{i,t,\ell}),\mathbf{e}^{r}\right).
\end{equation}
It measures how strongly this local region favors fake-related evidence over real-related evidence. To improve temporal stability, we aggregate the same patch location across the sampled frames:
\begin{equation}
u^{\mathrm{pat}}_{i,\ell}
=
\frac{1}{T_i}\sum_{t=1}^{T_i} g^{\mathrm{pat}}_{i,t,\ell}.
\end{equation}
Here, $u^{\mathrm{pat}}_{i,\ell}$ represents the cross-frame aggregated suspiciousness of the $\ell$-th patch location in video $x_i$. We do not use prototype concentration at the patch level because an individual patch usually contains only partial semantics, making concentration over prototypes substantially less reliable than at the whole-frame level.

Based on the resulting scores, we retain the top-$K_{\mathrm{frm}}$ frames according to $\{u^{\mathrm{frm}}_{i,t}\}$ and the top-$K_{\mathrm{pat}}$ patch locations according to $\{u^{\mathrm{pat}}_{i,\ell}\}$. The selected patch locations are then organized across the retained frames into cross-frame patch strips:
\begin{equation}
\mathcal{Z}_i=\{z_{i,m}\}_{m=1}^{K_{\mathrm{pat}}},
\end{equation}
where each $z_{i,m}$ is formed by the same selected patch location across the retained frames and serves as a localized evidence carrier for subsequent reasoning.

The resulting evidence set $\mathcal{Z}_i$ is then fed into Qwen together with a fixed instruction prompt $\rho$, which explicitly asks the model to describe localized forgery-related evidence rather than directly predicting whether the sample is real or fake. The resulting fine-grained textual description is written as:
\begin{equation}
q_i=\mathrm{Qwen}(\mathcal{Z}_i,\rho).
\end{equation}
Here, $q_i$ is treated as an explicit semantic representation of the mined suspicious evidence. To convert it into a comparable ranking signal, we parse $q_i$ into a set of evidence lines $\mathcal{L}(q_i)=\{\ell_{i,n}\}_{n=1}^{N_i}$, where $\ell_{i,n}$ denotes the $n$-th evidence line in $q_i$, and $N_i$ is the number of valid lines. We then compare each line against a fake anchor set $\mathcal{A}^{f}$ and a real anchor set $\mathcal{A}^{r}$ using a text reranker $h(\cdot,\cdot)$. The line-level margin is defined as:
\begin{equation}
m_{i,n}
=
\max_{a^{f}\in\mathcal{A}^{f}} h(\ell_{i,n},a^{f})
-
\max_{a^{r}\in\mathcal{A}^{r}} h(\ell_{i,n},a^{r}).
\end{equation}
The rank score $r_i$ is computed as the mean line-level margin. A larger $r_i$ indicates higher suspiciousness, and the corresponding video $x_i$ should therefore be assigned to a higher score slot within the uncertain subset.

\subsubsection{Slot Reordering within the Original Score Distribution}

As shown in Figure \ref{fig3}(3), the rank score $r_i$ is not used to define a new detector. Instead, it is used only to refine the local ordering of samples in the uncertain subset $\mathcal{U}$ while preserving the original score structure of the base detector.

To this end, we collect the original detector scores of all samples in $\mathcal{U}$ and sort them in descending order to form a set of fixed score slots, denoted by $\mathcal{V}_{\mathcal{U}}=\{v_k\}_{k=1}^{|\mathcal{U}|}$ with $v_1 \ge v_2 \ge \cdots \ge v_{|\mathcal{U}|}$. We then sort the uncertain samples according to their System-2 rank scores and reassign the slots in $\mathcal{V}_{\mathcal{U}}$ accordingly, so that samples with larger $r_i$ are mapped to higher-valued score slots. In this way, only the relative ordering within $\mathcal{U}$ is refined, while the slot values themselves remain unchanged. For samples in $\mathcal{C}$, the original scores are kept unchanged. For samples in $\mathcal{U}$, the fixed score slots in $\mathcal{V}_{\mathcal{U}}$ are reassigned according to the descending order of $r_i$, yielding the final refined scores $\hat{s}_i$.

By preserving the original uncertain-score slots and restricting refinement to the uncertain subset, this slot-preserving design maintains the original detector’s global decision structure while improving the relative ordering of hard samples whose ambiguity is not fully resolved by audio-visual assessment.

\begin{table*}[t]
\centering
% \small
\setlength{\tabcolsep}{16pt}
\renewcommand{\arraystretch}{1.150}

\definecolor{rowA}{RGB}{248,248,248}
\definecolor{rowB}{RGB}{239,239,239}
\definecolor{tfdsA}{RGB}{230,230,230}
\definecolor{tfdsB}{RGB}{220,220,220}
\definecolor{gainred}{RGB}{210,40,40}

\newcommand{\gain}[1]{\raisebox{-0.45ex}{\textcolor{gainred}{\scriptsize\textbf{#1}}}}

\begin{threeparttable}
\caption{Results on the AVLips, FKAV, and THB. We report AP (\%) and AUC (\%) with the best results in bold and the second-best results underlined on the full test sets. AVH-Align denotes the official checkpoint, and AVH-Align$^*$ denotes the retrained version under the AVLips training split. Red numbers indicate the absolute improvement brought by TFDS over its corresponding base detector. We use released checkpoints for supervised baselines. Since AVAD has no public training code, we evaluate it using its provided weights. }
\label{tab1}
\begin{tabular}{lcccccc}
\toprule
\multirow{2}{*}{\textbf{Methods}} 
& \multicolumn{2}{c}{\textbf{THB}} 
& \multicolumn{2}{c}{\textbf{AVLips}} 
& \multicolumn{2}{c}{\textbf{FKAV}} \\
\cmidrule(lr){2-3}\cmidrule(lr){4-5}\cmidrule(lr){6-7}
& \textbf{AP (\%)} & \textbf{AUC (\%)} & \textbf{AP (\%)} & \textbf{AUC (\%)} & \textbf{AP (\%)} & \textbf{AUC (\%)} \\
\midrule

\rowcolor{rowA}
CViT \cite{wodajo2021deepfake}
& 44.5 & 42.1 & 63.5 & 63.1 & 91.1 & 88.5 \\

\rowcolor{rowB}
EfficientViT \cite{coccomini2022combining}
& 31.6 & 21.7 & 63.3 & 64.8 & \textbf{95.1} & 90.9 \\

\rowcolor{rowA}
RealForensics \cite{haliassos2022leveraging}
& 68.7 & 74.3 & 69.9 & 71.9 & 94.2 & 88.2 \\

\rowcolor{rowB}
LipFD \cite{liu2024lips}
& 45.0 & 49.2 & 85.3 & 84.7 & 83.4 & 77.0 \\

\rowcolor{rowA}
AVAD \cite{feng2023self}
& 43.8 & 48.1 & 76.5 & 73.2 & 92.1 & 84.8 \\

\rowcolor{rowB}
AVH-Align \cite{smeu2025circumventing}
& 72.6 & 84.2 & 76.2 & 85.8 & 93.7 & 93.9 \\

\rowcolor{tfdsA}
AVH-Align+TFDS
& \textbf{77.5}\gain{+4.9} & \textbf{87.4}\gain{+3.2}
& \textbf{89.6}\gain{+13.4} & \underline{87.1}\gain{+1.3}
& \underline{95.0}\gain{+1.3} & \underline{94.2}\gain{+0.3} \\

\rowcolor{rowB}
AVH-Align$^{*}$ \cite{smeu2025circumventing}
& 64.8 & 82.3 & 74.3 & 84.5 & 93.5 & 93.0 \\

\rowcolor{tfdsB}
AVH-Align$^{*}$+TFDS
& \underline{77.0}\gain{+12.2} & \underline{87.3}\gain{+5.0}
& \underline{87.5}\gain{+13.2} & \textbf{89.7}\gain{+5.2}
& \textbf{95.1}\gain{+1.6} & \textbf{94.8}\gain{+1.8} \\

\bottomrule
\end{tabular}
\end{threeparttable}
\end{table*}

\section{Experiments}
\subsection{ Experimental Setup}
\subsubsection{Datasets and Metrics}
We conduct experiments on three talking head forgery detection datasets, including the publicly available AVLips \cite{liu2024lips}, FakeAVCeleb (FKAV) \cite{khalid2021fakeavceleb}, and TalkingHeadBench (THB) \cite{xiong2026talkingheadbench}. Among them, AVLips is split into training, validation, and test sets with a ratio of 6{:}1{:}3. The training set is used to retrain the self-supervised detector, whereas the validation split is used for detector selection during retraining and, after the detector is fixed, for threshold estimation in System-1. The resulting threshold is used not only to partition the AVLips test set into confident and uncertain samples, but also to perform the same partition on FKAV and THB, thereby ensuring a consistent routing criterion across different test sets. For FKAV, we construct the test set with 500 real samples and 1,000 selected fake samples. For THB, we merge the official test sets corresponding to videos generated by all diffusion models to form the target test set. Since THB does not contain real samples, we further supplement it with real videos from the AVLips test split. Following prior work, we use AUC and AP as the main evaluation metrics. AUC reflects the overall discriminative ability of the model, while AP evaluates performance from the precision-recall perspective. Together, they provide a comprehensive assessment of detection performance across different datasets.
\subsubsection{System-1 Detector Selection}
Our goal is not to compare the absolute performance of different self-supervised detectors, but to examine whether TFDS can further exploit the latent discriminative information of uncertain samples in a training-free manner on a detector that remains stable across datasets. Based on this consideration, we adopt the retrained AVH-Align \cite{smeu2025circumventing}, denoted by AVH-Align$^{*}$, as the default System-1 detector in the main experiments. We also consider another self-supervised detector, AVAD \cite{feng2023self}, but its performance collapses on some datasets. For example, both AUC and AP fall below 50\% on THB. It is therefore not suitable as a stable System-1 detector. In addition, we evaluate TFDS on top of the official AVH-Align checkpoint to examine whether its gain is sensitive to a particular initialization of the System-1 detector.
\subsubsection{ Implementation Details}
The comparative methods include both supervised and self-supervised talking head forgery detectors. We consider CViT \cite{wodajo2021deepfake}, EfficientViT \cite{coccomini2022combining}, RealForensics \cite{haliassos2022leveraging}, and LipFD \cite{liu2024lips} as supervised baselines, and AVAD \cite{feng2023self}, AVH-Align \cite{smeu2025circumventing} and AVH-Align$^{*}$ as self-supervised detectors. AVH-Align$^{*}$ is trained using the Adam optimizer on a single NVIDIA A100 GPU, with a learning rate of $9\times10^{-4}$ and a batch size of 1024. We use CLIP ViT-L/14 for evidence mining, Qwen2-VL-7B for fine-grained evidence description, and BGE-Reranker-Large for text-based rank score estimation. All predefined text prompts are generated by GPT-4. \textit{Additional implementation details and extended results are provided in the supplementary.}

\subsection{Experiment Results}
\subsubsection{Cross-Dataset Generalization}
Table \ref{tab1} reports the results on THB, AVLips, and FKAV. Under cross-dataset evaluation, AVH-Align shows more stable base performance than AVAD and outperforms most supervised detectors. This suggests that AVH-Align is less tied to specific forgery patterns or dataset biases, and is therefore better suited as the System-1 detector. By contrast, AVAD collapses on THB, where both AUC and AP fall below 50\%. A possible reason is that AVAD relies on generic audio-visual representations \cite{chen2021audio}, whereas AVH-Align is built on lip-reading-oriented representations \cite{shi2022learning} that are better aligned with talking head forgery detection. 

% Therefore, we choose AVH-Align and AVH-Align$^{*}$ as the base detector of System-1.

To assess whether TFDS depends on a particular detector initialization, we evaluate both the official AVH-Align checkpoint and a retrained version AVH-Align$^{*}$. The retrained model shows weaker base performance, which is expected given its more limited training data. TFDS consistently improves both AVH-Align versions across all datasets. On the official AVH-Align, it yields AP/AUC gains of +4.9/+3.2, +13.4/+1.3, and +1.3/+0.3 on THB, AVLips, and FKAV, respectively. On AVH-Align$^{*}$, the gains further increase to +12.2/+5.0, +13.2/+5.2, and +1.6/+1.8. This shows that TFDS is not tied to a single initialization, but remains effective across different initial coarse-ranking states. The larger gains on AVH-Align$^{*}$ can be attributed to its weaker initial score distribution, which leaves greater room for refining uncertain samples. Accordingly, we use AVH-Align$^{*}$ as the default System-1 detector in the subsequent analysis.

\begin{figure}[t]
\centering
\includegraphics[scale=0.43]{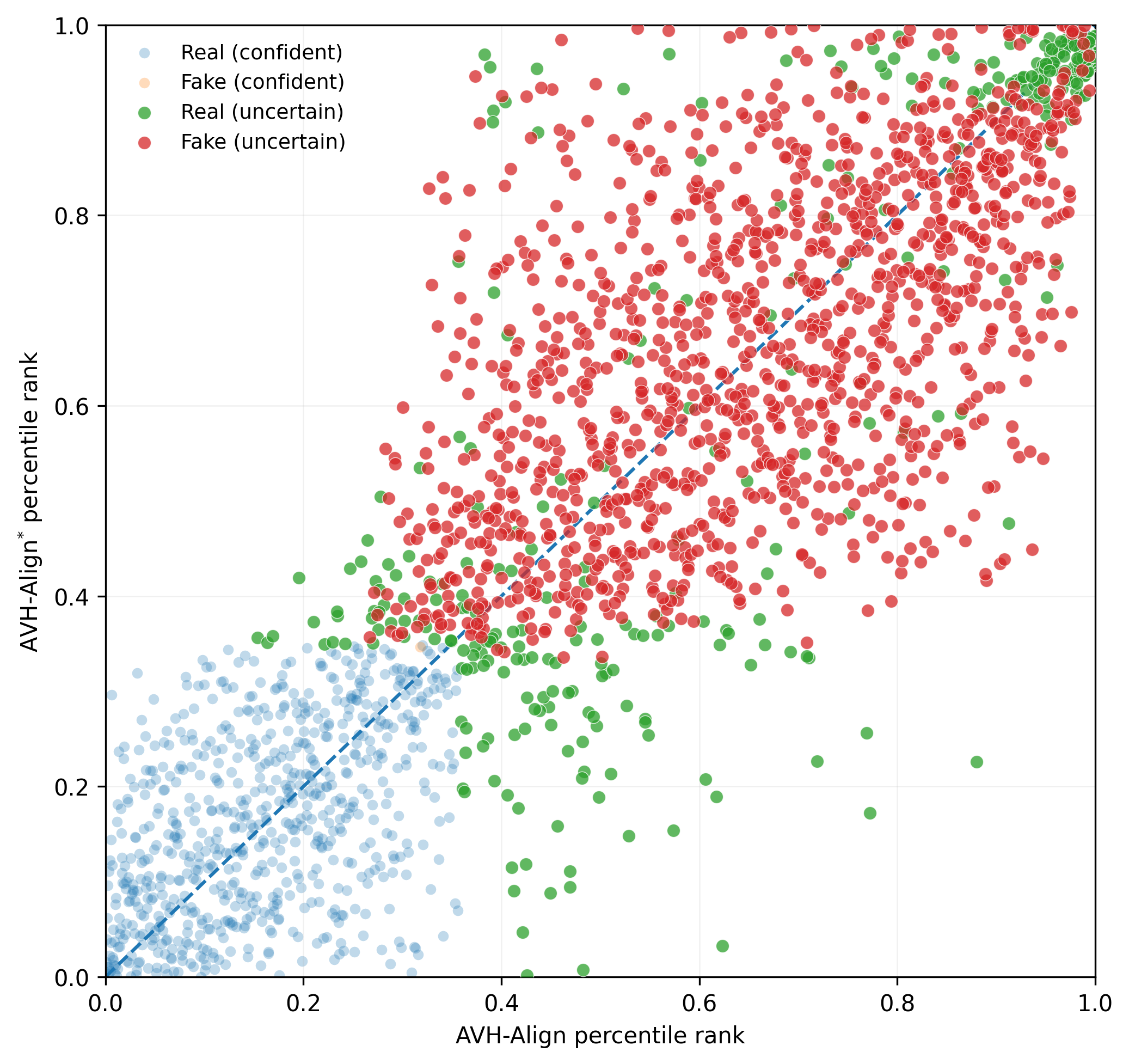}
\caption{Paired percentile ranks assigned by the official AVH-Align checkpoint and the retrained AVH-Align$^{*}$ on the AVLips test set. Each point denotes the same sample under the two detectors, with color indicating its real/fake label and confident/uncertain status.}
\label{fig4}
% \vspace{-0.5em}
\end{figure}
\subsubsection{Initial Score Distributions of System-1}
Figure \ref{fig4} compares the percentile ranks assigned by the official AVH-Align checkpoint and the retrained AVH-Align$^{*}$ on the AVLips test set. Although both detectors preserve a usable coarse-grained separation between real and fake samples, they induce noticeably different ranking structures. If the two detectors induced essentially the same ranking structure, most samples would lie near the diagonal. Instead, substantial deviations indicate markedly different relative positions for the same sample. This suggests that the official AVH-Align and AVH-Align$^{*}$ correspond to different initial coarse-ranking states.

Notably, these deviations are concentrated mainly in uncertain samples. This means that the main discrepancy between the two System-1 detectors does not lie in already well-separated easy samples, but in hard samples. In other words, the difference in initial score distributions is most pronounced precisely in the region where System-2 is intended to operate. It suggests that the gain of TFDS is not tied to a specific initial coarse-ranking state of the detector, since TFDS consistently improves both the official AVH-Align checkpoint and AVH-Align$^{*}$ despite their noticeably different initial ranking structures. 

\begin{table}[t]
\centering
% \small
\setlength{\tabcolsep}{5pt}
\renewcommand{\arraystretch}{1.20}

\definecolor{rowA}{RGB}{248,248,248}
\definecolor{rowB}{RGB}{239,239,239}
\definecolor{tfdsA}{RGB}{226,226,226}
\definecolor{gainred}{RGB}{210,40,40}

\newcommand{\gain}[1]{\raisebox{-0.45ex}{\textcolor{gainred}{\scriptsize\textbf{#1}}}}

\begin{threeparttable}
\caption{\textbf{Results on the uncertain subset.}
We report AP (\%) and AUC (\%) on the uncertain samples identified by System-1. AVH-Align$^{*}$ denotes the retrained AVH-Align under our training split. Red numbers indicate the absolute improvement brought by TFDS over its corresponding base detector.}
\label{tab2}
\begin{tabular}{lcccc}
\toprule
\multirow{2}{*}{\textbf{Methods}} 
& \multicolumn{2}{c}{\textbf{THB}} 
& \multicolumn{2}{c}{\textbf{AVLips}} \\
\cmidrule(lr){2-3}\cmidrule(lr){4-5}
& \textbf{AP (\%)} & \textbf{AUC (\%)} & \textbf{AP (\%)} & \textbf{AUC (\%)} \\
\midrule

\rowcolor{rowA}
AVH-Align$^{*}$
& 64.5 & 36.5 & 72.2 & 31.6 \\

\rowcolor{tfdsA}
AVH-Align$^{*}$+TFDS
& \textbf{77.1}\gain{+12.6} & \textbf{55.3}\gain{+18.8}
& \textbf{89.3}\gain{+17.1} & \textbf{67.1}\gain{+35.5} \\

\bottomrule
\end{tabular}
\end{threeparttable}
\end{table}

\subsubsection{Effectiveness on the Uncertain Subset} 
Table \ref{tab2} further reports the performance of TFDS on the uncertain subset. After integrating TFDS into AVH-Align$^{*}$, the AP/AUC on THB improves from 64.5/36.5 to 77.1/55.3, corresponding to gains of +12.6/+18.8. On AVLips, it improves from 72.2/31.6 to 89.3/67.1, yielding gains of +17.1/+35.5. These results show that, for the uncertain samples identified by System-1, System-2 can further enlarge their relative differences through fine-grained evidence mining and convert the latent discriminative information into more effective detection gains via local reordering. Moreover, the improvements on the uncertain subset are markedly larger than those on the full test set. This indicates that restricting refinement strictly to the uncertain subset not only preserves the overall structure of the original detector, but also more fully releases its remaining discriminative potential on hard samples.

\begin{figure}[t]
\centering
\includegraphics[scale=0.4]{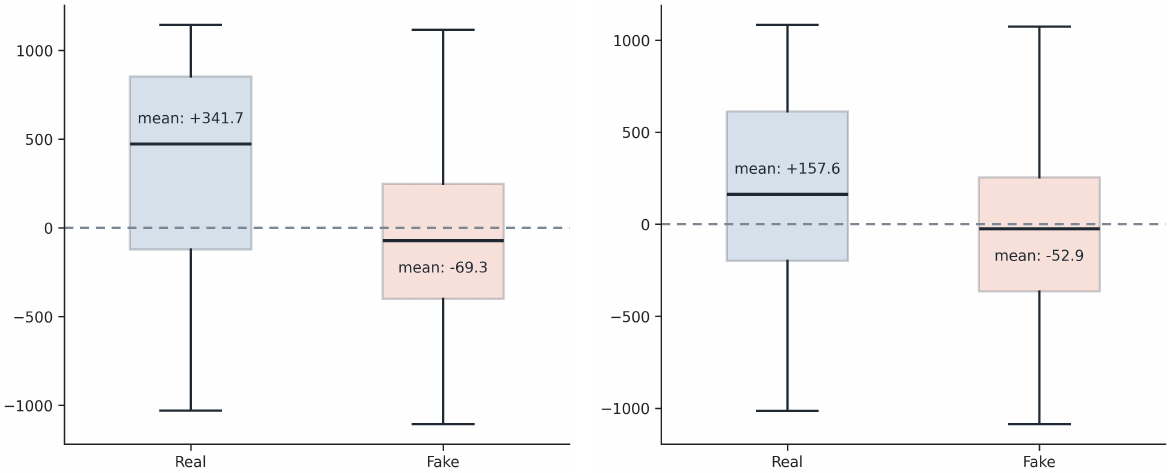}
\caption{Rank displacement of uncertain samples on AVLips (left) and THB (right). Negative values indicate that a sample is moved to a more suspicious position after refinement, while positive values indicate that it is moved to a less suspicious position. }
\label{fig5}
% \vspace{-0.5em}
\end{figure}

Furthermore, we quantify the rank displacement of uncertain samples before and after TFDS, as shown in Figure \ref{fig5}. Negative values indicate that a sample is moved to a more suspicious position after refinement, while positive values indicate that it is moved to a less suspicious position. On both AVLips and THB, real samples exhibit an overall positive shift, whereas fake samples show an overall negative shift. This indicates that TFDS tends to push more likely fake samples forward and move more likely real samples backward within the uncertain subset. This observation is consistent with the substantial gains reported in Table \ref{tab2}, and further suggests that the benefit of TFDS mainly comes from correcting the relative ordering of hard samples.

\begin{table*}[t]
\centering
% \small
\setlength{\tabcolsep}{9pt}
\renewcommand{\arraystretch}{1.20}

\definecolor{rowA}{RGB}{248,248,248}
\definecolor{rowB}{RGB}{239,239,239}
\definecolor{tfdsA}{RGB}{226,226,226}
\definecolor{gainred}{RGB}{210,40,40}

\newcommand{\gain}[1]{\raisebox{-0.45ex}{\textcolor{gainred}{\scriptsize\textbf{#1}}}}

\begin{threeparttable}
\caption{\textbf{Robustness under different perturbations on THB.} We report AP (\%) and AUC (\%) under inversion, noise, blur, and compression perturbations on the full test sets. AVH-Align$^{*}$ denotes the retrained AVH-Align under our training split. Red numbers indicate the absolute improvement brought by TFDS over its corresponding base detector.}
\label{tab3}
\begin{tabular}{lcccccccc}
\toprule
\multirow{2}{*}{\textbf{Methods}} 
& \multicolumn{2}{c}{\textbf{Inversion}} 
& \multicolumn{2}{c}{\textbf{Noise}} 
& \multicolumn{2}{c}{\textbf{Blur}} 
& \multicolumn{2}{c}{\textbf{Compress}} \\
\cmidrule(lr){2-3}\cmidrule(lr){4-5}\cmidrule(lr){6-7}\cmidrule(lr){8-9}
& \textbf{AP (\%)} & \textbf{AUC (\%)} 
& \textbf{AP (\%)} & \textbf{AUC (\%)} 
& \textbf{AP (\%)} & \textbf{AUC (\%)} 
& \textbf{AP (\%)} & \textbf{AUC (\%)} \\
\midrule

\rowcolor{rowA}
AVAD
& 31.0 & 20.9
& 40.1 & 40.3
& 39.3 & 39.3
& 38.6 & 38.0 \\

\rowcolor{rowB}
AVH-Align$^{*}$
& 41.5 & 44.0
& 46.3 & 53.0
& 37.4 & 36.2
& 38.4 & 37.9 \\

\rowcolor{tfdsA}
AVH-Align$^{*}$+TFDS
& \textbf{67.1}\gain{+25.6} & \textbf{72.5}\gain{+28.5}
& \textbf{67.3}\gain{+21.0} & \textbf{74.4}\gain{+21.4}
& \textbf{55.4}\gain{+18.0} & \textbf{50.8}\gain{+14.6}
& \textbf{55.9}\gain{+17.5} & \textbf{63.7}\gain{+25.8} \\

\bottomrule
\end{tabular}
\end{threeparttable}
\end{table*}

% required packages:
% \usepackage{booktabs}
% \usepackage[table]{xcolor}
% \usepackage{threeparttable}

\begin{table}[t]
\centering
% \small
\setlength{\tabcolsep}{3pt}
\renewcommand{\arraystretch}{1.20}

\definecolor{rowA}{RGB}{248,248,248}
\definecolor{rowB}{RGB}{239,239,239}
\definecolor{fullRow}{RGB}{226,226,226}

\begin{threeparttable}
\caption{\textbf{Ablation study on THB and AVLips.}  We report AP (\%) and AUC (\%) on the full test sets. The full model includes System-1 routing, CLIP-based frame and patch selection, Qwen-based evidence reasoning, reranker-based score estimation, and slot-preserving reordering. Best results are shown in \textbf{bold}.}
\label{tab4}
\begin{tabular}{lcccc}
\toprule
\multirow{2}{*}{\textbf{Method}} 
& \multicolumn{2}{c}{\textbf{THB}} 
& \multicolumn{2}{c}{\textbf{AVLips}} \\
\cmidrule(lr){2-3}\cmidrule(lr){4-5}
& \textbf{AP (\%)} & \textbf{AUC (\%)} & \textbf{AP (\%)} & \textbf{AUC (\%)} \\
\midrule

\rowcolor{rowA}
w/o System-1
& 48.9 & 46.0 & 50.0 & 54.6 \\

\rowcolor{rowB}
w/o CLIP-Frame Selector
& 72.8 & 85.6 & 86.1 & 88.9 \\

\rowcolor{rowA}
w/o CLIP-Patch Selector
& 71.9 & 85.1 & 82.8 & 87.0 \\

\rowcolor{rowB}
w/o CLIP
& 70.8 & 84.6 & 81.3 & 86.2 \\

\rowcolor{rowA}
w/o Qwen
& 69.6 & 83.8 & 75.7 & 83.5 \\

\rowcolor{rowB}
w/o Reranker
& 74.2 & 86.1 & 85.9 & 88.2 \\

\rowcolor{rowA}
w/o Slot
& 52.4 & 46.1 & 78.1 & 71.4 \\

\rowcolor{fullRow}
Full Model
& \textbf{77.0} & \textbf{87.3} & \textbf{87.5} & \textbf{89.7} \\

\bottomrule
\end{tabular}
\end{threeparttable}
\end{table}

\subsubsection{Robustness under Perturbations}
Table \ref{tab3} reports the detection results under inversion, noise, blur, and compression perturbations. All base detectors degrade substantially under these corruptions, indicating that talking head forgery detection remains highly challenging under out-of-distribution perturbations. Compared with AVAD, AVH-Align$^{*}$ maintains stronger base performance, but the base detector alone is still insufficient to handle complex perturbations reliably, especially under blur and compression. After TFDS is applied, this degradation is markedly alleviated. Relative to AVH-Align$^{*}$, TFDS yields AP/AUC gains of +25.6/+28.5, +21.0/+21.4, +18.0/+14.6, and +17.5/+25.8 under inversion, noise, blur, and compression, respectively.

These results show that the effectiveness of TFDS is not limited to clean data, but transfers consistently across multiple perturbation settings. Rather than relearning a new global decision boundary, System-2 further exploits residual discriminative information through fine-grained evidence mining and local reordering on samples that System-1 cannot handle reliably. 

% required packages:
% \usepackage{booktabs}
% \usepackage[table]{xcolor}
% \usepackage{threeparttable}

\subsection{Ablation Study}
To evaluate the contribution of each component, we conduct ablation experiments on THB and AVLips, as reported in Table \ref{tab4}.

\subsubsection{Effect of System-1 Routing}
Removing System-1 causes the most severe degradation, showing that System-2 should operate only on the uncertain subset rather than on all samples. In TFDS, System-1 is essential because it identifies where the original detector remains unreliable and where refinement is actually needed. Once this routing stage is removed, System-2 is applied indiscriminately to both hard and easy samples, which disrupts the reliable ordering already established for confident cases.  This result shows that the benefit of TFDS comes from targeted refinement on uncertain samples, rather than from applying System-2 as a global alternative to the base detector. 

\subsubsection{Effect of CLIP-based Frame Selection}
Replacing CLIP-based frame selection with fixed uniform frame sampling consistently degrades performance on both datasets. This shows that the gain of System-2 depends on whether the input frames already concentrate unresolved suspicious content. In the full model, frame selection uses global suspiciousness and prototype concentration to retain informative observations, so that subsequent reasoning is performed on evidence-rich frames rather than uniformly sampled ones.

\subsubsection{Effect of CLIP-based Patch Selection}
Removing CLIP-based patch selection and replacing it with uniform patch sampling causes a larger degradation than removing frame selection. This indicates that local evidence localization is more critical than temporal selection. Without patch selection, the visual evidence passed to Qwen becomes substantially less targeted. It demonstrates that fine-grained refinement should be driven by localized suspicious evidence rather than by frame-level inspection alone.

\subsubsection{Effect of Removing CLIP Guidance Entirely}
When CLIP guidance is removed, performance degrades further. This shows that the contribution of CLIP-based evidence mining comes from the joint effect of global and local selection. Frame selection first narrows the analysis to more informative visual observations, and patch selection then focuses reasoning on suspicious local regions.

\subsubsection{Effect of Qwen-based Evidence Reasoning}
Removing Qwen and directly using aggregated CLIP suspiciousness as the System-2 score leads to a pronounced drop. This confirms that coarse suspiciousness is insufficient to fully exploit the residual discriminative information from the base detector. Qwen plays the role of converting localized visual evidence into explicit fine-grained descriptions, thereby making subtle forgery cues more comparable across samples. Without this semantic reasoning stage, System-2 is reduced to score aggregation and can no longer perform fine-grained evidence disambiguation. 

\subsubsection{Effect of the Reranker}
Replacing the reranker with a simple keyword-count heuristic also degrades performance. This suggests that once Qwen has produced meaningful evidence descriptions, converting them into a stable ranking signal remains important. In the full model, the reranker aligns the generated evidence with real/fake text anchors in a more discriminative manner than a hand-designed keyword statistic. Its role is therefore not to replace reasoning, but to refine the mapping from semantic evidence to sortable scores. 

\subsubsection{Effect of Slot-Preserving Reordering}
Removing slot-preserving reordering causes one of the most severe performance drops. This shows that the System-2 score should not be used as a new global detector score, but only as a relative ranking signal within the uncertain subset. The slot-preserving mechanism enforces exactly this constraint by refining local ordering while preserving the original score structure of System-1. Once this constraint is removed, the refinement stage no longer respects the coarse but globally meaningful distribution produced by the base detector, and the overall detection structure is substantially damaged. This confirms that the benefit of TFDS lies in local correction under structural preservation, rather than in replacing the original detector with a new scoring function.

\section{Conclusion}

In this paper, we have proposed TFDS, a training-free dual-system framework for talking head forgery detection. Built on an existing self-supervised detector, TFDS first uses System-1 with score-based threshold estimation to partition test samples into confident and uncertain subsets. System-2 is then introduced to refine the uncertain samples through fine-grained evidence-guided reasoning and slot-preserving reordering. In this way, TFDS improves the ordering of ambiguous samples while preserving the original global score structure of the base detector. Extensive experiments show that TFDS consistently improves detection performance across multiple datasets and diverse perturbation settings. These results show that substantial gains can still be obtained from a fixed self-supervised detector by explicitly refining its uncertain predictions, without retraining a new detector.

% Overall, the ablation results support a coherent functional decomposition of TFDS. System-1 routing and slot-preserving reordering form the structural backbone of the framework, ensuring that refinement is activated only where needed and remains confined within the original score geometry. On top of this backbone, CLIP-based frame and patch selection improve the quality of the evidence passed to System-2, with local patch selection contributing more directly to fine-grained disambiguation. Qwen then converts localized suspicious regions into explicit semantic evidence, and the reranker transforms this evidence into a reliable ranking signal. Taken together, these results show that TFDS works not because of any isolated module, but because each component plays a distinct role in a coordinated uncertainty-aware refinement pipeline.

\bibliographystyle{ACM-Reference-Format}
\bibliography{sample-base}

@String{Computing = "Computing" }

@String{Computer = "{IEEE} Computer" }

@String{Springer = "Springer-Verlag" }

@ArtifactSoftware{R,
    title = {R: A Language and Environment for Statistical Computing},
    author = {{R Core Team}},
    organization = {R Foundation for Statistical Computing},
    address = {Vienna, Austria},
    year = {2019},
    url = {https://www.R-project.org/},
}

@article{liu2024lips,
  title={Lips are lying: Spotting the temporal inconsistency between audio and visual in lip-syncing deepfakes},
  author={Liu, Weifeng and She, Tianyi and Liu, Jiawei and Li, Boheng and Yao, Dongyu and Wang, Run},
  journal={Advances in Neural Information Processing Systems},
  volume={37},
  pages={91131--91155},
  year={2024}
}

@inproceedings{smeu2025circumventing,
  title={Circumventing shortcuts in audio-visual deepfake detection datasets with unsupervised learning},
  author={Smeu, Stefan and Boldisor, Dragos-Alexandru and Oneata, Dan and Oneata, Elisabeta},
  booktitle={Proceedings of the Computer Vision and Pattern Recognition Conference},
  pages={18815--18825},
  year={2025}
}

@article{khalid2021fakeavceleb,
  title={FakeAVCeleb: A novel audio-video multimodal deepfake dataset},
  author={Khalid, Hasam and Tariq, Shahroz and Kim, Minha and Woo, Simon S},
  journal={arXiv preprint arXiv:2108.05080},
  year={2021}
}

@inproceedings{peng2024synctalk,
  title={Synctalk: The devil is in the synchronization for talking head synthesis},
  author={Peng, Ziqiao and Hu, Wentao and Shi, Yue and Zhu, Xiangyu and Zhang, Xiaomei and Zhao, Hao and He, Jun and Liu, Hongyan and Fan, Zhaoxin},
  booktitle={Proceedings of the IEEE/CVF Conference on Computer Vision and Pattern Recognition},
  pages={666--676},
  year={2024}
}

@inproceedings{zheng2021exploring,
  title={Exploring temporal coherence for more general video face forgery detection},
  author={Zheng, Yinglin and Bao, Jianmin and Chen, Dong and Zeng, Ming and Wen, Fang},
  booktitle={Proceedings of the IEEE/CVF international conference on computer vision},
  pages={15044--15054},
  year={2021}
}

@inproceedings{feng2023self,
  title={Self-supervised video forensics by audio-visual anomaly detection},
  author={Feng, Chao and Chen, Ziyang and Owens, Andrew},
  booktitle={proceedings of the IEEE/CVF conference on computer vision and pattern recognition},
  pages={10491--10503},
  year={2023}
}

@article{shi2022learning,
  title={Learning audio-visual speech representation by masked multimodal cluster prediction},
  author={Shi, Bowen and Hsu, Wei-Ning and Lakhotia, Kushal and Mohamed, Abdelrahman},
  journal={arXiv preprint arXiv:2201.02184},
  year={2022}
}

@article{wei2023less,
  title={Less is better: Exponential loss for cross-modal matching},
  author={Wei, Jiwei and Yang, Yang and Xu, Xing and Song, Jingkuan and Wang, Guoqing and Shen, Heng Tao},
  journal={IEEE Transactions on Circuits and Systems for Video Technology},
  volume={33},
  number={9},
  pages={5271--5280},
  year={2023},
  publisher={IEEE}
}

@inproceedings{huang2023implicit,
  title={Implicit identity driven deepfake face swapping detection},
  author={Huang, Baojin and Wang, Zhongyuan and Yang, Jifan and Ai, Jiaxin and Zou, Qin and Wang, Qian and Ye, Dengpan},
  booktitle={Proceedings of the IEEE/CVF conference on computer vision and pattern recognition},
  pages={4490--4499},
  year={2023}
}

@article{yang2023avoid,
  title={Avoid-df: Audio-visual joint learning for detecting deepfake},
  author={Yang, Wenyuan and Zhou, Xiaoyu and Chen, Zhikai and Guo, Bofei and Ba, Zhongjie and Xia, Zhihua and Cao, Xiaochun and Ren, Kui},
  journal={IEEE Transactions on Information Forensics and Security},
  volume={18},
  pages={2015--2029},
  year={2023},
  publisher={IEEE}
}

@inproceedings{chugh2020not,
  title={Not made for each other-audio-visual dissonance-based deepfake detection and localization},
  author={Chugh, Komal and Gupta, Parul and Dhall, Abhinav and Subramanian, Ramanathan},
  booktitle={Proceedings of the 28th ACM international conference on multimedia},
  pages={439--447},
  year={2020}
}

@inproceedings{mittal2020emotions,
  title={Emotions don't lie: An audio-visual deepfake detection method using affective cues},
  author={Mittal, Trisha and Bhattacharya, Uttaran and Chandra, Rohan and Bera, Aniket and Manocha, Dinesh},
  booktitle={Proceedings of the 28th ACM international conference on multimedia},
  pages={2823--2832},
  year={2020}
}

@article{zeng2021contrastive,
  title={Contrastive learning of global and local video representations},
  author={Zeng, Zhaoyang and McDuff, Daniel and Song, Yale and others},
  journal={Advances in Neural Information Processing Systems},
  volume={34},
  pages={7025--7040},
  year={2021}
}

@inproceedings{haliassos2022leveraging,
  title={Leveraging real talking faces via self-supervision for robust forgery detection},
  author={Haliassos, Alexandros and Mira, Rodrigo and Petridis, Stavros and Pantic, Maja},
  booktitle={Proceedings of the IEEE/CVF conference on computer vision and pattern recognition},
  pages={14950--14962},
  year={2022}
}

@inproceedings{zhou2021joint,
  title={Joint audio-visual deepfake detection},
  author={Zhou, Yipin and Lim, Ser-Nam},
  booktitle={Proceedings of the IEEE/CVF international conference on computer vision},
  pages={14800--14809},
  year={2021}
}

@article{li2024zero,
  title={Zero-shot fake video detection by audio-visual consistency},
  author={Li, Xiaolou and Liu, Zehua and Chen, Chen and Li, Lantian and Guo, Li and Wang, Dong},
  journal={arXiv preprint arXiv:2406.07854},
  year={2024}
}

@inproceedings{ricker2024aeroblade,
  title={Aeroblade: Training-free detection of latent diffusion images using autoencoder reconstruction error},
  author={Ricker, Jonas and Lukovnikov, Denis and Fischer, Asja},
  booktitle={Proceedings of the IEEE/CVF Conference on Computer Vision and Pattern Recognition},
  pages={9130--9140},
  year={2024}
}

@inproceedings{coccomini2022combining,
  title={Combining efficientnet and vision transformers for video deepfake detection},
  author={Coccomini, Davide Alessandro and Messina, Nicola and Gennaro, Claudio and Falchi, Fabrizio},
  booktitle={International conference on image analysis and processing},
  pages={219--229},
  year={2022},
  organization={Springer}
}

@article{wodajo2021deepfake,
  title={Deepfake video detection using convolutional vision transformer},
  author={Wodajo, Deressa and Atnafu, Solomon},
  journal={arXiv preprint arXiv:2102.11126},
  year={2021}
}

@inproceedings{chencafe2025iclr,
  title={Cafe-Talk: Generating 3D Talking Face Animation with Multimodal Coarse-and Fine-grained Control},
  author={Chen, Hejia and Zhang, Haoxian and Zhang, Shoulong and Liu, Xiaoqiang and Zhuang, Sisi and Wan, Pengfei and ZHANG, Di and Li, Shuai and others},
  booktitle={The Thirteenth International Conference on Learning Representations},
  year={2025}
}

@article{qi2025global,
  title={Global prompt refinement with non-interfering attention masking for one-shot federated learning},
  author={Qi, Zhuang and Yu, Pan and Meng, Lei and Zhou, Sijin and Yu, Han and Li, Xiaoxiao and Meng, Xiangxu},
  journal={arXiv preprint arXiv:2509.22700},
  year={2025}
}

@article{zhou2022learning,
  title={Learning to prompt for vision-language models},
  author={Zhou, Kaiyang and Yang, Jingkang and Loy, Chen Change and Liu, Ziwei},
  journal={International journal of computer vision},
  volume={130},
  number={9},
  pages={2337--2348},
  year={2022},
  publisher={Springer}
}

@article{gao2024clip,
  title={Clip-adapter: Better vision-language models with feature adapters},
  author={Gao, Peng and Geng, Shijie and Zhang, Renrui and Ma, Teli and Fang, Rongyao and Zhang, Yongfeng and Li, Hongsheng and Qiao, Yu},
  journal={International journal of computer vision},
  volume={132},
  number={2},
  pages={581--595},
  year={2024},
  publisher={Springer}
}

@inproceedings{zhang2022tip,
  title={Tip-adapter: Training-free adaption of clip for few-shot classification},
  author={Zhang, Renrui and Zhang, Wei and Fang, Rongyao and Gao, Peng and Li, Kunchang and Dai, Jifeng and Qiao, Yu and Li, Hongsheng},
  booktitle={European conference on computer vision},
  pages={493--510},
  year={2022},
  organization={Springer}
}

@inproceedings{karmanov2024efficient,
  title={Efficient test-time adaptation of vision-language models},
  author={Karmanov, Adilbek and Guan, Dayan and Lu, Shijian and El Saddik, Abdulmotaleb and Xing, Eric},
  booktitle={Proceedings of the IEEE/CVF Conference on Computer Vision and Pattern Recognition},
  pages={14162--14171},
  year={2024}
}

@inproceedings{zhang2024dual,
  title={Dual memory networks: A versatile adaptation approach for vision-language models},
  author={Zhang, Yabin and Zhu, Wenjie and Tang, Hui and Ma, Zhiyuan and Zhou, Kaiyang and Zhang, Lei},
  booktitle={Proceedings of the IEEE/CVF conference on computer vision and pattern recognition},
  pages={28718--28728},
  year={2024}
}

@article{menon2022visual,
  title={Visual classification via description from large language models},
  author={Menon, Sachit and Vondrick, Carl},
  journal={arXiv preprint arXiv:2210.07183},
  year={2022}
}

@article{ren2023chatgpt,
  title={ChatGPT-powered hierarchical comparisons for image classification},
  author={Ren, Zhiyuan and Su, Yiyang and Liu, Xiaoming},
  journal={Advances in neural information processing systems},
  volume={36},
  pages={69706--69718},
  year={2023}
}

@inproceedings{pratt2023does,
  title={What does a platypus look like? generating customized prompts for zero-shot image classification},
  author={Pratt, Sarah and Covert, Ian and Liu, Rosanne and Farhadi, Ali},
  booktitle={Proceedings of the IEEE/CVF international conference on computer vision},
  pages={15691--15701},
  year={2023}
}

@inproceedings{qu2025proapo,
  title={Proapo: Progressively automatic prompt optimization for visual classification},
  author={Qu, Xiangyan and Gou, Gaopeng and Zhuang, Jiamin and Yu, Jing and Song, Kun and Wang, Qihao and Li, Yili and Xiong, Gang},
  booktitle={Proceedings of the Computer Vision and Pattern Recognition Conference},
  pages={25145--25155},
  year={2025}
}

@article{he2024rigid,
  title={Rigid: A training-free and model-agnostic framework for robust ai-generated image detection},
  author={He, Zhiyuan and Chen, Pin-Yu and Ho, Tsung-Yi},
  journal={arXiv preprint arXiv:2405.20112},
  year={2024}
}

@inproceedings{cozzolino2024zero,
  title={Zero-shot detection of ai-generated images},
  author={Cozzolino, Davide and Poggi, Giovanni and Nie{\ss}ner, Matthias and Verdoliva, Luisa},
  booktitle={European conference on computer vision},
  pages={54--72},
  year={2024},
  organization={Springer}
}

@article{tsai2024understanding,
  title={Understanding and improving training-free ai-generated image detections with vision foundation models},
  author={Tsai, Chung-Ting and Ko, Ching-Yun and Chung, I and Wang, Yu-Chiang Frank and Chen, Pin-Yu and others},
  journal={arXiv preprint arXiv:2411.19117},
  year={2024}
}

@article{choi2025training,
  title={Training-free Detection of AI-generated images via Cropping Robustness},
  author={Choi, Sungik and Lee, Hankook and Lee, Moontae},
  journal={arXiv preprint arXiv:2511.14030},
  year={2025}
}

@inproceedings{dong2025talking,
  title={Talking Head Generation via Viewpoint and Lighting Simulation Based on Global Representation},
  author={Dong, Biao and Zhang, Lei},
  booktitle={Proceedings of the 33rd ACM International Conference on Multimedia},
  pages={10258--10267},
  year={2025}
}

@inproceedings{yu2024gaussiantalker,
  title={Gaussiantalker: Speaker-specific talking head synthesis via 3d gaussian splatting},
  author={Yu, Hongyun and Qu, Zhan and Yu, Qihang and Chen, Jianchuan and Jiang, Zhonghua and Chen, Zhiwen and Zhang, Shengyu and Xu, Jimin and Wu, Fei and Lv, Chengfei and others},
  booktitle={Proceedings of the 32nd ACM International Conference on Multimedia},
  pages={3548--3557},
  year={2024}
}

@inproceedings{guo2025towards,
  title={Towards open-world generalized deepfake detection: General feature extraction via unsupervised domain adaptation},
  author={Guo, Midou and Yin, Qilin and Lu, Wei and Luo, Xiangyang},
  booktitle={Proceedings of the 33rd ACM International Conference on Multimedia},
  pages={11572--11580},
  year={2025}
}

@inproceedings{liu2023learning,
  title={Learning causality-inspired representation consistency for video anomaly detection},
  author={Liu, Yang and Xia, Zhaoyang and Zhao, Mengyang and Wei, Donglai and Wang, Yuzheng and Liu, Siao and Ju, Bobo and Fang, Gaoyun and Liu, Jing and Song, Liang},
  booktitle={Proceedings of the 31st ACM international conference on multimedia},
  pages={203--212},
  year={2023}
}

@inproceedings{kukanov2025klassify,
  title={KLASSify to Verify: Audio-Visual Deepfake Detection Using SSL-based Audio and Handcrafted Visual Features},
  author={Kukanov, Ivan and Ng, Jun Wah},
  booktitle={Proceedings of the 33rd ACM International Conference on Multimedia},
  pages={13707--13713},
  year={2025}
}

@article{zhang2025system,
  title={From system 1 to system 2: a survey of reasoning large language models},
  author={Zhang, Duzhen and Li, Zhong-Zhi and Zhang, Ming-Liang and Zhang, Jiaxin and Liu, Zengyan and Yao, Yuxuan and Xu, Haotian and Zheng, Junhao and Chen, Xiuyi and Zhang, Yingying and others},
  journal={IEEE Transactions on Pattern Analysis and Machine Intelligence},
  year={2025},
  publisher={IEEE}
}

@inproceedings{gu2025allm4add,
  title={Allm4add: Unlocking the capabilities of audio large language models for audio deepfake detection},
  author={Gu, Hao and Yi, Jiangyan and Wang, Chenglong and Tao, Jianhua and Lian, Zheng and He, Jiayi and Ren, Yong and Chen, Yujie and Wen, Zhengqi},
  booktitle={Proceedings of the 33rd ACM International Conference on Multimedia},
  pages={11736--11745},
  year={2025}
}

@inproceedings{huang2025sida,
  title={Sida: Social media image deepfake detection, localization and explanation with large multimodal model},
  author={Huang, Zhenglin and Hu, Jinwei and Li, Xiangtai and He, Yiwei and Zhao, Xingyu and Peng, Bei and Wu, Baoyuan and Huang, Xiaowei and Cheng, Guangliang},
  booktitle={Proceedings of the Computer Vision and Pattern Recognition Conference},
  pages={28831--28841},
  year={2025}
}

@inproceedings{yang2026endowing,
  title={Endowing Vision-Language Models with System 2 Thinking for Fine-grained Visual Recognition},
  author={Yang, Yutong and Huang, Lifu and Lin, Yijie and Peng, Xi and Yang, Mouxing},
  booktitle={Proceedings of the AAAI Conference on Artificial Intelligence},
  volume={40},
  number={14},
  pages={11802--11810},
  year={2026}
}

@article{yu2025unlocking,
  title={Unlocking the capabilities of large vision-language models for generalizable and explainable deepfake detection},
  author={Yu, Peipeng and Fei, Jianwei and Gao, Hui and Feng, Xuan and Xia, Zhihua and Chang, Chip Hong},
  journal={arXiv preprint arXiv:2503.14853},
  year={2025}
}

@article{fluss2005estimation,
  title={Estimation of the Youden Index and its associated cutoff point},
  author={Fluss, Ronen and Faraggi, David and Reiser, Benjamin},
  journal={Biometrical Journal: Journal of Mathematical Methods in Biosciences},
  volume={47},
  number={4},
  pages={458--472},
  year={2005},
  publisher={Wiley Online Library}
}

@inproceedings{radford2021learning,
  title={Learning transferable visual models from natural language supervision},
  author={Radford, Alec and Kim, Jong Wook and Hallacy, Chris and Ramesh, Aditya and Goh, Gabriel and Agarwal, Sandhini and Sastry, Girish and Askell, Amanda and Mishkin, Pamela and Clark, Jack and others},
  booktitle={International conference on machine learning},
  pages={8748--8763},
  year={2021},
  organization={PmLR}
}

@article{wang2024qwen2,
  title={Qwen2-vl: Enhancing vision-language model's perception of the world at any resolution},
  author={Wang, Peng and Bai, Shuai and Tan, Sinan and Wang, Shijie and Fan, Zhihao and Bai, Jinze and Chen, Keqin and Liu, Xuejing and Wang, Jialin and Ge, Wenbin and others},
  journal={arXiv preprint arXiv:2409.12191},
  year={2024}
}

@inproceedings{rachidy2025domain,
  title={Domain Adaptive Document Reranking for Retrieval Augmented Generation},
  author={Rachidy, Yassine and Hmamouche, Youssef and Sehbaoui, Faissal and Seghrouchni, Amal El Fallah},
  booktitle={2025 IEEE 37th International Conference on Tools with Artificial Intelligence (ICTAI)},
  pages={1381--1387},
  year={2025},
  organization={IEEE}
}

@inproceedings{li2024llms,
  title={LLMs as bridges: Reformulating grounded multimodal named entity recognition},
  author={Li, Jinyuan and Li, Han and Sun, Di and Wang, Jiahao and Zhang, Wenkun and Wang, Zan and Pan, Gang},
  booktitle={Findings of the Association for Computational Linguistics: ACL 2024},
  pages={1302--1318},
  year={2024}
}

@inproceedings{oh2025understanding,
  title={Understanding Multimodal LLMs Under Distribution Shifts: An Information-Theoretic Approach},
  author={Oh, Changdae and Fang, Zhen and Im, Shawn and Du, Xuefeng and Li, Yixuan},
  booktitle={International Conference on Machine Learning},
  pages={46943--46970},
  year={2025},
  organization={PMLR}
}

@inproceedings{wang2024lampmark,
  title={Lampmark: Proactive deepfake detection via training-free landmark perceptual watermarks},
  author={Wang, Tianyi and Huang, Mengxiao and Cheng, Harry and Zhang, Xiao and Shen, Zhiqi},
  booktitle={Proceedings of the 32nd ACM International Conference on Multimedia},
  pages={10515--10524},
  year={2024}
}

@article{brown2020language,
  title={Language models are few-shot learners},
  author={Brown, Tom and Mann, Benjamin and Ryder, Nick and Subbiah, Melanie and Kaplan, Jared D and Dhariwal, Prafulla and Neelakantan, Arvind and Shyam, Pranav and Sastry, Girish and Askell, Amanda and others},
  journal={Advances in neural information processing systems},
  volume={33},
  pages={1877--1901},
  year={2020}
}

@article{sanderson2023gpt,
  title={GPT-4 is here: what scientists think},
  author={Sanderson, Katharine},
  journal={Nature},
  volume={615},
  number={7954},
  pages={773},
  year={2023},
  publisher={Springer Science and Business Media LLC}
}

@inproceedings{xiong2026talkingheadbench,
  title={Talkingheadbench: A multi-modal benchmark \& analysis of talking-head deepfake detection},
  author={Xiong, Xinqi and Patel, Prakrut and Fan, Qingyuan and Wadhwa, Amisha and Selvam, Sarathy and Guo, Xiao and Qi, Luchao and Liu, Xiaoming and Sengupta, Roni},
  booktitle={Proceedings of the IEEE/CVF Winter Conference on Applications of Computer Vision},
  pages={4139--4149},
  year={2026}
}

@article{chen2021audio,
  title={Audio-visual synchronisation in the wild},
  author={Chen, Honglie and Xie, Weidi and Afouras, Triantafyllos and Nagrani, Arsha and Vedaldi, Andrea and Zisserman, Andrew},
  journal={arXiv preprint arXiv:2112.04432},
  year={2021}
}

\end{document}